\pdfoutput=1

\documentclass[11pt]{article}

\usepackage{acl}

\usepackage{times}
\usepackage{latexsym}

\usepackage[T1]{fontenc}

\usepackage[utf8]{inputenc}

\usepackage{microtype}

\usepackage{amsmath}
\usepackage{amssymb}
\usepackage{amsfonts}
\usepackage{csquotes}
\usepackage{fancyhdr}
\usepackage{float}
\usepackage{graphicx}

\fancypagestyle{FirstPage}{
\cfoot{Copyright \copyright~2020 Association for Computational Linguists (ACL). All Rights Reserved.}
}

\title{Error correction and extraction in request dialogs}

\author{Stefan Constantin \and Alex Waibel \\
  Karlsruhe Institute of Technology \\
  Institute for Anthropomatics and Robotics \\
  \texttt{$\{$stefan.constantin|waibel$\}$@kit.edu}}

\begin{document}
\maketitle
\begin{abstract}
We propose a dialog system utility component that gets the last two utterances of a user and can detect whether the last utterance is an error correction of the second last utterance.
If yes, it corrects the second last utterance according to the error correction in the last utterance and outputs the extracted pairs of reparandum and repair entity.
This component offers two advantages, learning the concept of corrections to avoid collecting corrections for every new domain and extracting reparandum and repair pairs, which offers the possibility to learn out of it.

For the error correction one sequence labeling and two sequence to sequence approaches are presented.
For the error correction detection these three error correction approaches can also be used and in addition, we present a sequence classification approach.
One error correction detection and one error correction approach can be combined to a pipeline or the error correction approaches can be trained and used end-to-end to avoid two components.
We modified the EPIC-KITCHENS-100 dataset to evaluate the approaches for correcting entity phrases in request dialogs.
For error correction detection and correction, we got an accuracy of 96.40\,\% on synthetic validation data and an accuracy of 77.81\,\% on human-created real-world test data.
\end{abstract}

\section{Introduction}
\thispagestyle{FirstPage}
Errors and ambiguities are difficult to avoid in a dialog.
Corrections allow to recover from errors and to disambiguate ambiguities.
For example, a household robot gets the request \enquote{Put the cleaned spoons into the cutlery drawer}, but the robot does not know which one of the drawers is the cutlery drawer.
It can choose one of the drawers and put the spoons there.
If its choice is wrong, the user must correct the robot, e.\,g. \enquote{No, into the drawer right of the sink}.
Alternatively, the robot can ask which one of the drawers is the cutlery drawer.
The clarification response of the user, e.\,g. \enquote{It's the drawer right of the sink}, is also a correction because the response disambiguates the ambiguity.
Another type of correction occurs when users change their mind, e.\,g. \enquote{I changed my mind, the forks}, or when the system misunderstands the user request (e.\,g. because of automatic speech recognition or natural language understanding errors).

All these correction types can be processed in the same manner and therefore we propose a component that gets a request and a correction and outputs a corrected request.
To get this corrected request, the phrases in the correction phrase replace their corresponding phrases in the request.
In this work, we restrict on entity phrases like \enquote{drawer right of the sink}.
To replace other phrases like verb phrases is out of scope for this work.
The request \enquote{Put the cleaned spoons into the cutlery drawer} with its correction \enquote{No, into the drawer right of the sink} is converted to \enquote{Put the cleaned spoons into the drawer right of the sink}.
Such a component has two advantages compared to handling the corrections in the actual dialog component.
First, it reduces the amount of required training data for the actual dialog component because corrections will not need to be learned if there is an open-domain correction component.
Second, this kind of correction component can be extended so that it outputs the extracted pairs of reparandum and repair entity.
In our example there is one pair: \enquote{cutlery drawer} and \enquote{drawer right of the sink}.
These entity pairs can be used, for example, for learning in a life-long learning component of a dialog system to reduce the need for correction in future dialogs, e.\,g. the robot can learn which one of the drawers is the cutlery drawer.

\section{Related Work}
Studies have been conducted in the area of interactive repair dialog.
In \cite{SuhmMW1996} a multi-modal approach is used.
The user can highlight wrong phrases and respeak or spell the correct phrase, or choose from alternatives in the n-best list of the automatic speech recognition component, or use handwriting to write the correct phrase.
These error strategies are improved in \cite{SuhmW1997} by considering the context.
In \cite{SuhmMW1999,SuhmMW2001} the previous approaches are evaluated in more detail in a dictation system with real users.
Different human strategies for error correction are presented in \cite{Gieselmann2006}.

\citet{SagawaMN2004} propose an error handling component based on correction grammars.
These correction grammars have the advantage that they can be used domain-independently.
However, they need a grammar-based dialog system.
An error correction detection module and strategies to handle the detected errors are proposed by \citet{GriolM2016}.
The corrected request must be handled by the Spoken Language Understanding component.
That means, for every domain the Spoken Language Understanding component must be adapted to the possible corrections.
\citet{KraljevskiH2017} propose a domain-independent correction detection by checking the speech for hyperarticulation.
Other features than hyperarticulation are not used.

In \cite{BechetF2013}, a system is presented that detects errors in automatic speech recognition transcripts and asks the user for a correction.

There are also studies that research automatic error correction without user interaction.
In \cite{XieAAJN2016}, a character-based approach to correct language errors is used.
They used a character-based approach to avoid out-of-vocabulary words because of orthographic errors.
In \cite{WenMK...2020}, the authors used a multi-task setup to correct the automatic speech recognition outputs and do the natural language understanding.

The task of request correction presented in the introduction is related to the task of disfluency removal.
In disfluency removal, there are the reparandum (which entity should be replaced), the interruption point (where the correction begins), the interregnum (which phrase is the signal phrase for the correction), and the repair phrase (the correct entity) \cite{Shriberg1994}.

In Figure \ref{fig:DisfluencyTerminology}, a disfluent utterance annotated with this terminology is depicted.

\begin{figure}[H]
\centering

spoon into the \(\underbrace{\text{drawer}}_{\text{reparandum}}\) \(\underbrace{\text{}}_{\text{interruption pt.}}\) \(\underbrace{\text{uh}}_{\text{interregnum}}\) \(\underbrace{\text{sink}}_{\text{repair}}\)

\caption{disfluent utterance annotated with repair terminology}
\label{fig:DisfluencyTerminology}
\end{figure}

A lot of work has been conducted for disfluency removal \cite{ChoNW2014,DongWYXX2019,WangCL2016, JamshidAJ2018}.
In all these works, it is assumed that it is enough to delete tokens of the disfluent utterance to get a fluent utterance.
A disfluent utterance with the copy and delete labels is depicted in Figure \ref{fig:DisfluencyRemoval}.

\begin{figure}[H]
\centering

\begin{tabular}{c c c c c c}
spoons & into & the & drawer & uh & sink \\
C & C & C & D & D & C
\end{tabular}

\caption{disfluent utterance labeled with copy and delete labels}
\label{fig:DisfluencyRemoval}
\end{figure}

However, in the task of corrections, long-distance replacements can occur.
That means, that between the reparandum and the repair are words that are important and must not be deleted.
Such a long-distance replacement is depicted in Figure \ref{fig:CorrectionTerminology}.

\begin{figure}[H]
\centering

\[ \underbrace{\text{spoon}}_{\text{reparandum}} \text{into the drawer} \underbrace{\text{}}_{\text{interruption pt.}} \underbrace{\text{no}}_{\text{interregnum}}
\underbrace{\text{forks}}_{\text{repair}} \]

\caption{request and correction phrase annotated with repair terminology}
\label{fig:CorrectionTerminology}
\end{figure}

\section{Dataset}
\label{sec:dataset}
Our dataset is based on the annotations in natural language of the EPIC-KITCHENS-100 dataset \cite{DamenDF2020Collection, DamenDF2020Rescaling}.
The EPIC-KITCHENS-100 dataset comprises 100 hours of recordings of actions in a kitchen environment.
An example annotation of such an action is \enquote{put pizza slice into container} and the corresponding verb is \enquote{put-into} and the corresponding entities are \enquote{slice:pizza} and \enquote{container}.
Annotations in this dataset have one verb and zero to six entities.
The verb, the corresponding verb class, the entities and the corresponding entity classes are explicitly saved to every annotation.
The order of the entities and the corresponding entity classes is the same as in the annotation.
If the verb has a preposition, the verb is saved including its preposition.
The words of the entities are represented in a hierarchy.
The most general word of the hierarchy is left and the words are more specific the further they are to the right of the hierarchy.
The words of each hierarchy are separated by a colon.
There are 67\,218 annotations in the training and 9669 annotations in the validation dataset of the EPIC-KITCHENS-100 dataset.
There is no test dataset.
Some annotations occur multiple times, because different recordings of the 100 hours recordings have the same annotation.
By considering only the unique annotations, 15\,968 annotations are in the training and 3835 annotations are in the validation dataset.

For our dataset, we used only the annotations that have one or two entities.
We excluded the annotations with no entities because we need at least one entity that can be corrected.
Annotations including more than two entities amount only to less than 1.15\,\% of all annotations and therefore we decided to exclude them because of dataset balancing reasons.

\begin{figure}
\includegraphics[trim=1.5cm 8cm 1cm 10cm,clip,width=0.5\textwidth]{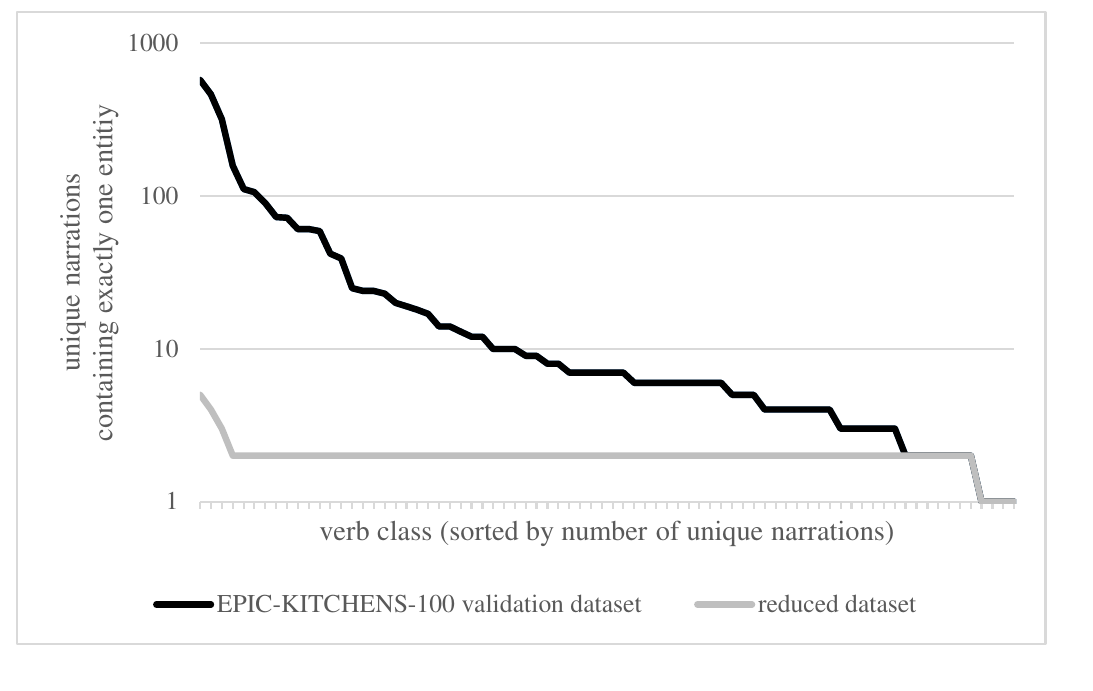}

\includegraphics[trim=1.5cm 8cm 1cm 10cm,clip,width=0.5\textwidth]{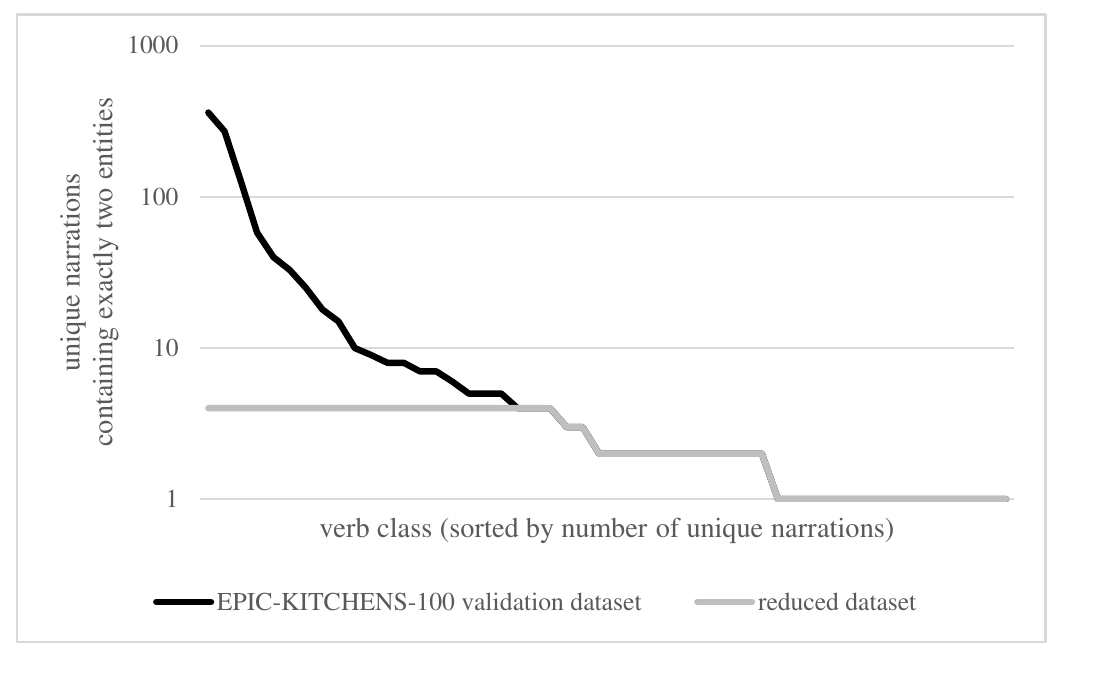}
\caption{unique annotations in respect to the verb class before and after the reduction of the EPIC-KITCHENS-100 validation dataset}
\label{fig:statistics}
\end{figure}

The verb classes of the EPIC-KITCHENS-100 datasets are imbalanced.
To get a better balance in the validation dataset, we removed annotations of verb classes that occur very often from the validation dataset.
We wanted a more balanced dataset to evaluate whether the model gets along with very different verb classes.
We calculated the number of desired remaining annotations of a verb class, called $r$, by dividing the number of annotations, called $a$, by 100, but we determined a minimal number of remaining annotations of verb classes: 2 for one entity annotations ($r = max(2, a/100)$) and 4 for two entity annotations ($r = max(4, a/100)$).
In some cases, there are less than the desired remaining annotations of a verb class in the EPIC-KITCHENS-100 dataset then we used the possible number.
We chose the values for minimal examples to get a nearly balanced dataset: 142 annotations with one entity and 122 annotations with two entities.
To get the annotations of a verb class, we chose the verbs occurring in a verb class equally distributed.
In total, we have 264 annotations in the reduced validation dataset.
The numbers of unique annotations in respect to the verb class before and after the  reduction are depicted in Figure \ref{fig:statistics}.

In the EPIC-KITCHENS-100 dataset, the training and validation datasets are similar: all 78 verb classes of the validation dataset occur in the training dataset and 346 of the 372 first level word of the entity hierarchies of the validation dataset occur in the training dataset.
Because of this, we decided to reduce the training dataset to have more difference between them.
We removed the verb classes of the 49 less frequent occurring verb classes (in total 98 verb classes are in the training dataset) from the training dataset and removed all entities from the training dataset if its first part was also in the validation dataset.
That means, if bowl:washing:up was in the validation dataset, an annotation with bowl:salad in the training dataset was removed.
After the reduction 4822 annotations were left in the training dataset.

To use these annotations for training and evaluating the error correction detection and correction component, we had to add corrections to the annotations.
For the training and validation dataset, we generated the corrections synthetically.
There are three options for the entity replacement: the first entity should be replaced, the second entity should be replaced, or both entities should be replaced.
We drew uniformly distributed which of these three options should be applied.
If both entities should be corrected, we drew uniformly distributed in which order they should be corrected.
For the training and validation dataset, we had 8 and 6, respectively, templates to introduce the correction phrase, followed by the corrected entities.
An entity could be replaced by an entity that occurs in an annotation of the same verb class in the same position.
An example for one corrected entity is \enquote{Be so kind and pick the oregano} for the request and \enquote{it's the chilli} for the correction and an example for two corrected entities is \enquote{Could you put the tin in the Cupboard?} for the request and \enquote{no the olives in the Fridge} for the correction.

For the test dataset, we had nine human data collectors who could freely write the corrections, they only knew what entities should be replaced with what other entities (but they were allowed to use synonyms for the other entities) and whether the correction should be a correction to a wrong action of the robot, a clarification, or a correction because the user changed their mind (equally distributed).

We added 19 and 14 templates before the narration to increase the variety of the natural language of the training and validation dataset, respectively.
In the EPIC-KITCHENS-100 dataset, the articles of the entities are missing, therefore we added a \enquote{the} before the entities.
For the test dataset, we used the narrations of our validation dataset and let the same nine annotators that created the corrections for our test dataset paraphrase them.

The test dataset is more challenging than the validation dataset because it differs even more from the training dataset.
The nine data collectors were told to use a large variety of natural language.

We used the 4822 annotations of the reduced training dataset to generate with the different data augmentations 52\,357 request and correction pairs for the error correction training dataset. The error correction validation dataset has 264 request and correction pairs and the error correction test dataset has 960 request and correction pairs.

To train and evaluate the error correction detection, we need examples where the last utterance is no correction.
To achieve this, the second last and the last utterance are made of all the requests of the error correction data.
The requests were shuffled for the last utterance.
This approach doubled the number of examples to the correction examples, that means, we have 104\,714 pairs in the error correction detection and error correction training dataset, 528 pairs in the error correction detection and error correction validation dataset, and 1920 pairs in the error correction detection and error correction test dataset.

The target for the error correction datasets is the corrected request and the reparandum repair pairs and the target for the error correction detection and error correction dataset depends whether the source has a request and correction pair or a request and request pair.
In the first case, there is an error correction and the target is the same as in the error correction datasets, in the second case, the target is to copy both requests.
There is a further dataset, the error correction detection dataset.
The sources are the same as in the error correction detection and error correction dataset but the target is the binary value whether there is a correction or not.

We created the described datasets in different forms for the different approaches.
For the sequence labeling approach, we labeled the source tokens with different labels, see Figure \ref{fig:ExampleDataSeqLabeling} and Section \ref{sec:models} for an explanation of the labels.
 
For the sequence to sequence approach with generative token generation, we created source and target pairs, see Figure \ref{fig:ExampleDataSeq2Seq}.
For the sequence to sequence approach with generation by copying source tokens, we added the order of copy operations.
Additionally, the separator tokens that are needed in the target will be inserted to the source, see Figure \ref{fig:ExampleDataSeq2SeqCopy}.

\begin{figure}[H]
would C \\
it C \\
be C \\
possible C \\
to C \\
wash C \\
the C \\
table R1 \\
? C \\
| D \\
no D \\
the D \\
wok S1 \\
instead D \\
of D \\
the D \\
table D \\
. D
\caption{sequence labeling data example}
\label{fig:ExampleDataSeqLabeling}
\end{figure}

\begin{figure}[H]
source file:
Would it be possible to wash the table ? | no the Wok instead of the table .

target file:
Would it be possible to wash the Wok ? | table -> Wok
\caption{sequence to sequence with fixed vocabulary data example}
\label{fig:ExampleDataSeq2Seq}
\end{figure}

\begin{figure}[H]
source file:
Would it be possible to wash the table ? | no the Wok instead of the table . - -> 

target file:
Would it be possible to wash the Wok ? | table -> Wok

copy target file (considering the T5 prefix and the T5 tokenization):
3 4 5 6 7 8 9 16 17 11 12 13 10 26 27 16 17 28
\caption{sequence to sequence with copy source token approach data example}
\label{fig:ExampleDataSeq2SeqCopy}
\end{figure}

\section{Models}
\label{sec:models}
For the error correction and extraction, we developed three different approaches.
The first approach is a sequence labeling approach, the second approach is a sequence to sequence approach where the output tokens are sampled from a fixed vocabulary, and the third approach is a sequence to sequence approach where output tokens are copied from the source tokens.

For the sequence labeling approach, every word is labeled with one of the following labels: C (copy), D (delete), R1 (entity 1, potentially to be replaced), R2 (entity 2, potentially to be replaced), S1 (entity to replace entity 1), or S2 (entity to replace entity 2).
For the correction target, the S1 and S2 labeled entities are used to replace the R1 and R2 labeled entities, respectively.
For the extraction target, the output is the pairs R1 and S1 as well as R2 and S2 if there is a replacement available for the first or second entity, respectively.
In Figure \ref{fig:Approach}, an example request and correction pair is labeled and both targets are given.

\begin{figure*}
\centering
\setlength{\tabcolsep}{4pt}
\begin{tabular}{ | p{3.2cm} | c c c c c c || c c c c c c c c | }
\hline
request \textbar\textbar~\mbox{correction} & put & the & milk & into & the & shelf & no & the & soja & milk & into & the & left & shelf \\
\hline
labels & C & R1 & R1 & R2 & R2 & R2 & D & S1 & S1 & S1 & S2 & S2 & S2 & S2 \\
\hline
corrected request & \multicolumn{14}{c |}{put the soja milk into the left shelf} \\
\hline
pairs of reparandum and repair entity & \multicolumn{14}{c |}{milk \(\rightarrow\)~
soja milk -~into the shelf \(\rightarrow\)~into the left shelf} \\
\hline
\end{tabular}

\caption{error correction example}
\label{fig:Approach}
\end{figure*}

For the sequence labeling, we propose fine-tuning the cased BERT large model (24 Transformer encoder blocks, hidden size of 1024, 16 self-attention heads, and 340 million parameters) \cite{DevlinCLT2019}.

For the sequence to sequence approach where the output tokens are sampled from a fixed vocabulary, we propose fine-tuning a T5 large model \cite{RaffelSR...2020}.
The T5 model is a pre-trained Transformer network \cite{VaswaniSPUJGKP2017} and the T5 large model has the following properties: 24 Transformer encoder blocks, 24 Transformer decoder blocks, hidden size of 1024 (in- and output) and 4096 (inner-layer), 16 self-attention heads, 737 million parameters.

The probability distribution over the fixed vocabulary \(V\) can be calculated in the following way:
\[P_{\text{generate}}(V) = softmax(dec^T \cdot W_{\text{generate}})\]
where \(dec\) is the output of the Transformer decoder and \(W_{\text{generate}} \in \mathbb{R}^{\text{hidden size decoder} \times \text{vocabulary size}}\) is a learnable matrix.

We call this T5 model T5 generate.

In the corrected request there are only tokens of the input sequence.
To utilize this property, we developed a pointer network model \cite{VinyalsFJ2015} with the T5 large model that calculates which input token has the highest probability to be copied to the output sequence.
This is our third approach.
The probability distribution over the input sequence tokens \(V'\) can be calculated in the following way:
\[P_{\text{copy}}(V') = softmax(dec^T \cdot enc^T)\]
where \(dec\) is the output of the Transformer decoder and \(enc \in \mathbb{R}^{\text{source input length} \times \text{embedding size}}\).

To utilize the knowledge of the pre-trained model, we feed the source input token with the highest probability into the encoder instead of the position of the source input token.
That means, that in the generation stage the copy mechanism is only used, otherwise it is like a normal T5 model.
To be able to output the separators, we add these to the source, so that they can also be copied.
We call this modificated T5 model T5 copy.

To decide whether an utterance is a correction for the previous request command, the described three approaches can also be used.
If all output labels of the sequence labeling approach are C, no error correction is detected, otherwise there is an error correction.
The sequence to sequence approaches detect an error correction if the source and the target without the separators are not equal, otherwise there is no error correction.
In the T5 copy approach, the source for the comparison is the original source and not the source with the inserted separators.

In addition to these three approaches, a sequence classification can also be used for the error correction detection.
For the sequence classification, we propose to fine-tune the cased BERT large model (24 Transformer encoder blocks, hidden size of 1024, 16 self-attention heads, and 340 million parameters) \cite{DevlinCLT2019}.

\section{Implementation}
We used the HuggingFace \cite{WolfDS...2020} Pytorch \cite{PaszkeGM...2019} BERT and T5 models for our implementations of the models described in Section \ref{sec:models} and published our implementations and our models~\footnote{\url{https://github.com/msc42/seq2seq-transformer} \url{https://github.com/msc42/seq-labeling-and-classification}}.

\section{Evaluation}
\begin{table*}
\centering

\begin{tabular}{ p{0.9cm} p{2.1cm} | p{1.25cm} | p{1.25cm} p{1.25cm} p{1.32cm} | p{1.25cm} p{1.25cm} p{1.32cm} }
 & & & \multicolumn{3}{c}{detecting corrections} & \multicolumn{3}{c}{detecting no corrections} \\
dataset & model & accuracy & precision & recall & F\textsubscript{1}-score & precision & recall & F\textsubscript{1}-score \\ 
\hline
valid. & classification & 100\,\% & 100\,\% & 100\,\% & 100\,\% & 100\,\% & 100\,\% & 100\,\% \\
valid. & seq. labeling & 100\,\% & 100\,\% & 100\,\% & 100\,\% & 100\,\% & 100\,\% & 100\,\% \\
valid. & T5 generate & 98.67\,\% & 98.13\,\% & 99.24\,\% & 98.68\,\% & 99.23\,\% & 98.11\,\% & 98.67\,\% \\
valid. & T5 copy & 71.78\,\% & 63.92\,\% & 100\,\% & 77.99\,\% & 100\,\% & 43.56\,\% & 60.69\,\% \\
\hline
test & classification & 87.86\,\% & 99.86\,\% & 75.83\,\% & 86.20\,\% & 80.52\,\% & 99.90\,\% & 89.17\,\% \\
test & seq. labeling & 88.49\,\% & 99.87\,\% & 77.08\,\% & 87.01\,\% & 81.34\,\% & 99.90\,\% & 89.67\,\% \\
test & T5 generate & 84.01\,\% & 96.58\,\% & 70.52\,\% & 81.52\,\% & 76.78\,\% & 97.50\,\% & 85.91\,\% \\
test & T5 copy & 77.45\,\% & 73.63\,\% & 85.52\,\% & 79.13\,\% & 82.73\,\% & 69.38\,\% & 75.47\,\% \\
\end{tabular}

\caption{evaluation results of the error correction detection, all models except the classification were trained on the error correction detection and error correction dataset and the classification was trained on the error correction detection dataset}
\label{tab:ResultsErrorDetection}
\end{table*}

\begin{table*}
\centering

\begin{tabular}{l| l l l | l l l }
 & \multicolumn{3}{c}{validation dataset} & \multicolumn{3}{c}{test dataset} \\
model & correction & extraction & both & correction & extraction & both \\ 
\hline
seq. labeling & 96.21\,\% & 94.70\,\% & 94.70\,\% & 40.10\,\% & 48.75\,\% & 39.06\,\% \\
E2E seq. labeling & 96.59\,\% & 95.08\,\% & 95.08\,\% & 39.27\,\% & 43.54\,\% & 38.65\,\% \\
T5 generate & 92.80\,\% & 95.83\,\% & 91.29\,\% & 73.65\,\% & 77.81\,\% & 71.98\,\% \\
E2E T5 generate & 96.21\,\% & 95.08\,\% & 94.70\,\% & 37.40\,\% & 38.75\,\% & 36.25\,\% \\
T5 copy & 50.38\,\% & 87.12\,\% & 50.00\,\% & 50.52\,\% & 62.19\,\% & 47.92\,\% \\
E2E T5 copy & 70.83\,\% & 92.42\,\% & 68.94\,\% & 27.50\,\% & 35.00\,\% & 25.31\,\% \\
\end{tabular}

\caption{evaluation results of the error correction (metric accuracy), the end-to-end (E2E) models were trained on the error correction detection and error correction dataset and the other models were trained on the error correction dataset}
\label{tab:ResultsErrorCorrection}
\end{table*}

\begin{table*}
\centering

\begin{tabular}{p{5.4cm} | p{1.2cm} p{1.2cm} p{1.05cm} | p{1.2cm} p{1.2cm} p{1.05cm} }
model(s) & \multicolumn{3}{c}{validation dataset} & \multicolumn{3}{c}{test dataset} \\
  & correction & extraction & both & correction & extraction & both \\ 
\hline
detection and seq. labeling & 98.11\,\% & 97.35\,\% & 97.35\,\% & 67.08\,\% & 68.44\,\% & 66.72\,\% \\
detection and E2E seq. labeling & 98.30\,\% & 97.54\,\% & 97.54\,\% & 69.58\,\% & 71.77\,\% & 69.27\,\% \\
E2E seq. labeling & 98.30\,\% & 97.54\,\% & 97.54\,\% & 69.58\,\% & 71.77\,\% & 69.27\,\% \\
\hline
detection and T5 generate & 96.40\,\% & 98.67\,\% & 96.40\,\% & 78.49\,\% & 79.84\,\% & 77.81\,\% \\
detection and E2E T5 generate & 98.11\,\% & 98.30\,\% & 98.11\,\% & 68.33\,\% & 68.80\,\% & 67.66\,\% \\
E2E T5 generate & 97.54\,\% & 97.16\,\% & 96.40\,\% & 68.07\,\% & 68.49\,\% & 66.88\,\% \\
\hline
detection and T5 copy & 75.19\,\% & 94.13\,\% & 75.00\,\% & 68.44\,\% & 72.71\,\% & 66.93\,\% \\
detection and E2E T5 copy & 85.42\,\% & 96.59\,\% & 84.66\,\% & 63.28\,\% & 66.77\,\% & 62.08\,\% \\
E2E T5 copy & 69.70\,\% & 78.03\,\% & 56.25\,\% & 55.00\,\% & 58.91\,\% & 47.40\,\% \\
\end{tabular}

\caption{evaluation results of the error correction detection and error correction (metric accuracy), the end-to-end (E2E) models were trained on the error correction detection and error correction dataset and the other models were trained on the error correction dataset, \enquote{and} means that the error correction detection was done by the best error correction detection model (sequence labeling) and the error correction detection by the model mentioned after the \enquote{and} if a correction was detected}
\label{tab:ResultsErrorDetectionAndCorrection}
\end{table*}

In this section, we will first evaluate the different error correction detection component approaches described in Section \ref{sec:models}.
After that, the error correction component approaches described in Section \ref{sec:models} are evaluated.
Third, we will compare whether it is better to separate the error correction detection and error correction in separate components and use a pipeline approach or whether an end-to-end approach is better.
For all evaluations, we used the datasets described in Section \ref{sec:dataset}.
 	
We fine-tuned the sequence classification and labeling approaches one epoch with the following hyperparameters: AdamW optimizer \cite{LoshchilovH2019} with learning rate of \(2 \cdot 10^{-5}\), batch size of 32 and maximum input length of 128.

The T5 generate and T5 copy models were fine-tuned one epoch with the following hyperparameters: Adam optimizer \cite{KingmaB2014} with learning rate of \(2.5 \cdot 10^{-4}\), batch size of 24 and a maximum input length of 128; in the embedding layer, the first two encoder blocks were frozen.

The results of the error correction detection components are depicted in Table \ref{tab:ResultsErrorDetection}.
Accuracy means how many examples were classified correctly, precision is how many of the positive classified examples are really positive, recall how many of the positive examples are found by the component and the F\textsubscript{1}-score is the harmonic mean of the precision and recall.
We calculated the precision, recall and F\textsubscript{1}-score for the case that detecting corrections were the positive examples and for the case that detecting no corrections were the positive examples to get better insights in the quality of the differently trained models.
The sequence classification approach was trained with the error correction detection dataset and the other approaches were trained with the error correction detection and error correction dataset.
The best approach is the sequence labeling approach (if all words have the copy label C, it is no error correction, otherwise it is an error correction).
It has an accuracy of 100\,\% for the validation and 88.49\,\% for the test dataset.
The recall for detecting no corrections is 99.90\,\% and the precision 81.34\,\% (F\textsubscript{1}-score 89.67\,\%) in the test dataset.
That means, if there is no correction, the component detects it in most of the cases and make no unnecessary correction.
This is a good property, because it is better not detecting a correction than correcting something which is already correct.
The error correction detection and error correction component should improve the overall system and not make it worse.
Nevertheless, the results for detecting corrections with a recall of 77.08\,\% and a precision of 99.87\,\% (F\textsubscript{1}-score 87.01\,\%) in the test dataset are good.
In some cases where the component fails, it is really difficult to detect the correction like in \enquote{Kindly turn off the heat on the oven | Please turn off the water tap on the oven}.
The classification approach has similar results to the sequence labeling approach: 100\,\% accuracy for the validation dataset and 87.86\,\% for the test dataset.
This approach also prefers detecting no corrections over corrections.
The T5 generate approach is worse.
It has an accuracy of 98.67\,\% on the validation dataset and an accuracy of 84.01\,\% on the test dataset.
The worst results are from the T5 copy approach (71.78\,\% and 77.45\,\% validation and test dataset accuracy, respectively).

The results of the error correction components are depicted in Table \ref{tab:ResultsErrorCorrection}.
We evaluated the error correction with the metric accuracy.
The correction is correct if the predicted correction and the reference correction are the same.
The extraction of the reparandum and repair pairs is correct if the predicted pairs are equal to the reference pairs. The order and entities that map to themselves are ignored.
Both are correct if the correction as well as the extraction are correct.
For this evaluation the error correction datasets are used.
On the validation dataset, the sequence labeling approach that was trained on the error correction detection and error correction dataset has the best overall accuracy (95.08\,\%).
The accuracy for the correction is 96.59\,\% and for the extraction 95.08\,\%.
On the test dataset, the T5 generate approach that is trained on the error correction dataset has the best accuracy (71.98\,\%).
In general, all approaches trained on the error correction detection and error correction dataset have a higher accuracy on the validation dataset and all approaches trained on the error correction dataset have a higher accuracy on the test dataset.
The T5 copy extraction could be optimized by bookkeeping the order of copy operations, stopping after finishing the correction and use the bookkeeping to reconstruct the reparandum and repair pairs.
We relinquished this optimization because the correction results were much worse and we did not see any sense in further optimizations that will only lead to minimal improvements.

The results of the error correction detection and error correction components are depicted in Table~\ref{tab:ResultsErrorDetectionAndCorrection}.
We used the same metric accuracy as in the error correction evaluation.
For the error correction detection in the pipeline approach, we used the best error correction detection model evaluated in this section.
It is the sequence labeling approach where no correction is in the example if all labels are C.
After the error correction detection, the error correction will occur.
We evaluated all three approaches described in Section \ref{sec:models} in their version trained on the error correction detection and error correction dataset and their version trained on the error correction dataset.
In the end-to-end setting, a component executes the error correction detection and the error correction in one run.
The best approach is the pipeline approach with the T5 generate approach only trained on the error correction dataset with an accuracy of 96.40\,\% on the validation and 77.81\,\% accuracy on the test dataset.

The evaluation results show that the test dataset is more challenging than the validation dataset.
The nine data collectors were able to introduce even more variety of natural language than the validation dataset has.

\section{Conclusions and Further Work}
The	 proposed error correction detection and error correction component shows high potential.
For the validation dataset, we got very good results: in 96.40\,\% of the cases, we could detect whether there is a correction or not and if there is a correction, it outputs a correct corrected request and could extract correctly the reparandum and repair pairs.
The results for the human-generated real-world data with 77.81\,\% shows that the proposed component is learning the concept of corrections and can be developed to be used as an upstream component to avoid the need for collecting data for request corrections for every new domain.
In addition, the extraction of the pairs of reparandum and repair entity can be used for learning in a life-long learning component of a dialog system to reduce the need for correction in future dialogs.

In future work, the training dataset could be extended to a bigger variety of natural language which will enable the model to learn the concept of corrections better and to get better results on human-generated real-world data.
The mentioned life-long learning component could also be part of future work and the classification of correction types could improve the performance of such a life-long learning component.
To improve the accuracy, architectures that have a better NER performance than our used BERT model, like the architecture proposed by \cite{BaevskiELZA2019}, could be used.
A further future research goal is to be able to correct all phrases and not only entity phrases.

\section*{Acknowledgements}
This work has been supported by the German Federal Ministry of Education and Research (BMBF) under the project OML (01IS18040A).

\bibliography{custom}

\end{document}